\RequirePackage{fix-cm}
\documentclass[smallextended]{svjour3}       
\smartqed  
\usepackage{graphicx}
\usepackage{amsthm}
\usepackage{amsmath}
\usepackage{array}
\usepackage{amssymb}
\usepackage{enumitem}

\theoremstyle{definition}
\newtheorem{myDef}{Definition}
\newtheorem{myExamp}{Example}
\newtheorem{mytheorem}{Theorem}  
  
\newtheorem*{myProof}{Proof}
\newtheorem{myExprm}{Experiment}
\newtheorem{myProc}{Procedure}

\usepackage{algorithm}
\usepackage{algpseudocode}

\usepackage{etoolbox}
\newcommand*{\affaddr}[1][*]{\textsuperscript{#1}}
\newcommand*{\affmark}[1][*]{\textsuperscript{#1}}
\usepackage[misc]{ifsym}

\begin{document}

\title{Attributed Rhetorical Structure Grammar for Domain Text Summarization}


\author{Ruqian Lu\affmark[1,2] \and Shengluan Hou\affmark[2,3] \and Chuanqing Wang\affmark[1,3] \and Yu Huang\affmark[2,3] \and Chaoqun Fei\affmark[2,3] \and Songmao Zhang\affmark[1,2]}

\institute{\Letter \quad Ruqian Lu \at
           rqlu@math.ac.cn
           \and \affaddr[1] Key Laboratory of MADIS, Academy of Mathematics and Systems Science), Chinese Academy of Sciences, Beijing 100190, China
           \\
           \at \affaddr[2] Key Laboratory of Intelligent Information Processing, Institute of Computing Technology, Chinese Academy of Sciences, Beijing 100190, China
           \\
           \at \affaddr[3] University of Chinese Academy of Sciences, Beijing 100049, China
}

\date{Received: date / Accepted: date}

\maketitle

\begin{abstract}This paper presents a new approach of automatic text summarization which combines domain oriented text analysis (DoTA) and rhetorical structure theory (RST) in a grammar form: the attributed rhetorical structure grammar (ARSG), where the non-terminal symbols are domain keywords, called domain relations, while the rhetorical relations serve as attributes. We developed machine learning algorithms for learning such a grammar from a corpus of sample domain texts, as well as parsing algorithms for the learned grammar, together with adjustable text summarization algorithms for generating domain specific summaries. Our practical experiments have shown that with support of domain knowledge the drawback of missing very large training data set can be effectively compensated. We have also shown that the knowledge based approach may be made more powerful by introducing grammar parsing and RST as inference engine. For checking the feasibility of model transfer, we introduced a technique for mapping a grammar from one domain to others with acceptable cost. We have also made a comprehensive comparison of our approach with some others.

\end{abstract}

\section{Introduction}
\label{Introduction}

Summarizing information and knowledge from NL texts without relying on NLU techniques has always been a challenging task for the information engineers. Fortunately, we have seen the emerging and development of successful linguistic theories which support the analyzing, understanding and summarizing of NL texts.

There are usually two different approaches for text summarization \cite{gambhir2017recent,nenkova2011automatic}: the abstractive approach \cite{DurrettBK16,YoshidaSHN14,rush2015neural,cao2018retrieve} and the extractive counterpart~\cite{lei2018An,Marcu2000The,Barzilay1997,hirao2013single,SilberM02,ErcanC08,cheng2016neural,zhou2018neural}.
In these works various techniques of machine learning have been proposed to challenge the difficulties of automatic text summarization. It is often that machine learning has to be done over hundreds of thousands articles \cite{cheng2016neural,zhou2018neural}, or even millions of such samples \cite{rush2015neural,cao2018retrieve}. The quantity and quality of sample articles to be learned from may affect the capability and efficiency of the final summarizer essentially. This problem is even more serious if the information one wants from text summarization is domain oriented. We may have a huge set of news articles in general. But it is difficult to have access to a huge set of articles in a specialized area. This is even worse regarding domain oriented Chinese texts summarization due to the lack of large sized open sources.

One way out of this difficulty is to take the (domain) knowledge based approach. In this approach, the lack or scarce of domain oriented training data may be compensated by pre-collected domain knowledge of the summarizer. With well-organized domain knowledge even a limited set of training data may produce good quality summarizers. For example, most lexical chain based methods leverage extern knowledge base (KB) to construct lexical chains and use them as lexical cohesion
features for text summarization \cite{Barzilay1997,SilberM02,ErcanC08,pourvali2012automated}, Genest and Lapalme used a knowledge base to identify patterns in the representation of the source documents and generate summary text from them\cite{genest2013absum}. Timofeyev and Choi first derived syntactic structure and mapped words to Cyc. By clustering mapped Cyc concepts, main topics were identified and then used for identifying informative concepts and forming the summary sentences.

However, although these works make use of concepts as representation of knowledge, they do not make use of inference as driving engine of knowledge. This weakness has affected the result of summarization. 
In this paper we propose a knowledge based, grammar supported, rhetorical structure theory (RST) \cite{Mann1988Rhetorical} enriched approach for automatic text summarization, where grammatical inference and RST serve as knowledge driving engine. 

Rhetorical Structure Theory (RST) was formally proposed as a technique of studying discourse structure of natural language texts~\cite{Marcu2000The,Mann1988Rhetorical,CarlsonMO01}. The use of RST in automatic generation of text summarization was a bit later. It was Daniel Marcu who has finished the first work on practical RST based discourse parsing and text summarization~\cite{Marcu97}. Central to RST is the notion of coherence, so that RST can also be understood as a theory of text coherence. The basic components of RST are the (rhetorical) relations (RRE for short). RRE holds between two sibling nodes under a common parent node, where one child node is nucleus (N) with the remaining one a satellite (S). It is possible that both children are nuclei. In terms of the writer's purpose, what expressed in N is more essential than what expressed in S. N is comprehension independent of S, but not vice versa.

We have designed a framework of attribute grammar where both terminal and non-terminal symbols are concepts of a given domain. For each production rule the rule head is an abstraction or consequence of the rule body. The rhetorical relations serve as attributes of non-terminal domain concepts. Other attributes include symbolic markers such as the cues and punctuations. The grammar is non-deterministic due to the complexity of natural language representation of domain texts. In fact it is probabilistic where the probability comes from machine learning on training texts. Besides, the grammar is oriented towards precedential bottom up parsing. In the following sections we will provide details of how such a grammar is learned from a set of training articles and how it is applied to summarizing domain texts automatically. We call the resulting model an Attributed Rhetorical Structure Grammar (ARSG).

The remaining part of this paper is organized as follows:  Section~\ref{RelatedWorks} is about related works. Section~\ref{BasicConceptsOfOurApproachAndARSG} introduces the basic concepts of our approach and ARSG. Section~\ref{GrammarLearning} is about grammar learning and parser implementation. Section~\ref{ARSGBasedTS} is about ARSG based text summarization and its optimization. Section~\ref{ExperimentalResults} presents experimental results. Section \ref{ModelTransduction} introduces a technique of model transduction. Section \ref{ApproachComparation} compares our approach with two representative approaches. Concluding remarks are given in Section~\ref{ConcludingRemarks}.

\section{Related works}
\label{RelatedWorks}
Classical extractive summarization approaches rely on textual features for summary generation, which include RST based, lexical chain based, as well as other machine learning based methods.

The RST approaches for single document summarization have got good results~\cite{DurrettBK16,YoshidaSHN14,Marcu2000The}. Generally they first construct a rhetorical structure tree (RTR) and then select the content based on the RTR~\cite{Marcu2000The,HiraoYNYN13}, where the core is RST analysis which may be based on rules, machine learning techniques or deep learning networks. Marcu et al. were the first to use rule-based methods for discourse analysis by benefiting information derived from a corpus study of tree phrases~\cite{Marcu2000The,Marcu97}. LeThanh et al. used syntactic textual operations to generate best discourse structures~\cite{Lethanh2004Generating}. Tofiloski et al. presented a discourse segmenter (SLSeg) leading to higher precision~\cite{TofiloskiBT09}. Machine learning approaches utilize probabilistic models, SVM and CRF. Soricut and Marcu used probabilistic models SPADE for sentence segmentation and tree building to extract features from the syntactic tree~\cite{SoricutM03}. 

Many researches focused on SVM-based discourse analysis. They regarded relation identification as classification problem~\cite{HernaultPdI10,FengH12}. Joty et al. first used Dynamic Conditional Random Field (DCRF) for sentence-level discourse analysis~\cite{JotyCNM13}, following HILDA~\cite{HernaultPdI10}. Recent advances in deep learning led to further progress in NLP~\cite{Julia2015Advances}. Ji and Eisenstein's representation learning based method~\cite{JiE14Rep} is the state-of-art method in identifying relation types. However, deep learning method has poor interpretability. On the other hand, the current RST analyzers are insufficient for practical use.

Lexical chain exploits the lexical cohesion among related words and is a mixed syntactic and semantic approach~\cite{MorrisH91}, which has been widely used in text summarization~\cite{Barzilay1997,SilberM02,ErcanC08} and some other NLP areas~\cite{GalleyM03,WeiLCZB15,QianJZTX14}, because of its easy computation and high efficiency. One of its main approaches focuses on the use of knowledge resources, such as WordNet~\cite{Miller95}. Hirst and St-Onge presented the first implementation based on greedy disambiguation~\cite{Hirst1998lexical}. Barzilay and Elhadad proposed a new method for word sense disambiguation (WSD) \cite{Barzilay1997}, but with exponential complexity. Galley and McKeown gained best results by separating WSD from lexical chain construction~\cite{GalleyM03}. Remus and Biemann used the Latent Dirichlet allocation (LDA) topic model for estimating semantic closeness by interpreting lexical chains as clusters~\cite{RemusB13}. Barzilay presented a method of scoring chains as summary sentences~\cite{Barzilay1997}. Ercan represented topics by sets of co-located lexical chains to get the final summary~\cite{ErcanC08}.

Graph-based methods are another type of classical methods where each text is represented as a graph, whose nodes and edges correspond to textual units (often sentences or words) and their similarity respectively. According to the basic idea of PageRank~\cite{page1998pagerank}, these models rank and select the top-$n$ sentences (words) as the final summary~\cite{Mihalcea2004TextRank}. Utilizing statistical tools, Fattah proposed a hybrid machine learning model based on maximum entropy, naive-Bayes, and SVM model as well as textual features to improve summary content selection~\cite{fattah2014hybrid}. For more details, see~\cite{gambhir2017recent,NenkovaM12}.

With large available datasets and high performance computing devices, deep neural networks have shown great promise in text summarization. Rush et al. first applied attention framework to abstractive text summarization on large scale datasets (Gigaword\footnote{https://catalog.ldc.upenn.edu/ldc2003t05}, DUC2004\footnote{https://www-nlpir.nist.gov/projects/duc/data/2004\_data.html}) with good performance~\cite{rush2015neural}. For further advancements see~\cite{cao2018retrieve,chopra2016abstractive,nallapati2016abstractive}. Cheng and Lapata proposed a deep extractive model, including a hierarchical document reader and an attention-based content extractor~\cite{cheng2016neural}, trained and tested on huge datasets DailyMail~\cite{hermann2015teaching} and DUC 2002, where the attention is directly applied to decoder rather than to hidden units~\cite{see2017get}. Their improvements include SummaRuNNer of Nallapati et al.~\cite{nallapati2017summarunner},  and NEUSUM of Zhou et al.~\cite{zhou2018neural} which achieved the best result on CNN/DailyMail dataset.

\section{Basic Concepts and Attributed Rhetorical Structure Grammar}
\label{BasicConceptsOfOurApproachAndARSG}

\begin{myDef}
A domain knowledge base (DKB) is composed of three kinds of concepts:

1. The acting agents of a domain (called green concepts);

2. The major influence factors of a domain (called red concepts);

3. The concepts about dynamics of a domain (called blue concepts).\hfill $\square$\par
\label{threeDictDef}
\end{myDef}

Note that the third kind of domain concept (DCP) is also called domain relation (DRE). Each kind of concepts forms a hierarchy ordered in different levels. A domain concept is not necessary a single word. It may also be a phrase.
\begin{myExamp}
The domain of World Economy and Trade (WET):

1. Green concepts: ASEAN, USA, developing countries, BRICS, IMF, Bank of China, etc;

2. Red concepts: market, price, inflation, trade, policy, GDP, PMI, stock, future, option, etc;

3. Blue concepts:  stabilized rise, turbulent, gladness with worry, spurt, soared, etc.
\label{wetDictExamp}
\end{myExamp}

\begin{myDef}
A lexical core, LC for short, is a triple of \{green, red, blue\} concepts. In the parsing process of ARSG to be defined later they appear as the three nodes of a hight-1 binary tree called basic tree, where blue concept is the root and {green, red} concepts are the two leaves.
\hfill $\square$\par
\label{LCDef}
\end{myDef}

Note the difference between lexical core and lexical chain~\cite{Barzilay1997,SilberM02}. Traditionally a lexical chain connects words of similar meaning and often runs through the whole text. It doesn't reflect the meaning of a single sentence unit because of its non-locality (not at clause level), and its non-conformity (consisting of unstructured single words with incomplete information). On the contrary, a lexical core is a part of a sentence or clause (locality), characterizing the domain relatedness (conformity) of its elements. More exactly, properties of DRE include:

1. Each DRE denotes the change or not change or the way of change of some state of that domain;

2. Domains themselves are organized in hierarchies, where each child domain inherits its parent domain's DRE in addition to its own.

Throughout this paper we will use the following text as example of grammar parsing and content summarization. 

\begin{myExamp}
A text from the domain WET:

\noindent
(C1) It is well known that although China's foreign trade develops rapidly and (C2) China's integration into the global value chain is increasing, (C3) China is still in a lower position in the international division of labor. (C4) Especially in the high-tech industries, technology and services exports and so on, (C5) China is still at the low end of the global value chain. (C6) Therefore, how to speed up the transformation and upgrading of foreign trade, (C7) and to enhance the status of international division of labor in China, (C8) is an important factor of China's economic development in the future.
\label{ftTextExamp}
\end{myExamp}

\begin{myExamp}
From the clause C1 we extract the LC:
\begin{center}
\{China (green), foreign trade (red), develops rapidly (blue)\}
\end{center}
\label{extractLCExamp}
\end{myExamp}

Attribute grammar was invented based on the idea that a perfect grammar parsing should not be only based on its syntax, but also on its semantics~\cite{Knuth68}. That means the parsing process should `bring' some information with it. Such information is called attributes, therefore the name of attribute grammar. In top down (bottom up) parsing the attributes are calculated and brought downwards (upwards) to terminals (root) of the grammar. They are called inherited (synthesized) attributes. The grammars using these attributes are called l-attributed (s-attributed) grammars. As a working tool for text summarization, we prefer the s-attributed grammar.

We differentiate between syntactic attributes and semantic attributes. The former is those syntactic elements of NL texts, which have influence on ARSG parsing, e.g. the cues, punctuations, segmentations, and even paragraph hierarchies. The latter characterizes the role of DRE and rhetorical relation (RRE) during parsing. It performs two functions, of which the first one is doing value calculation and transmission. This is traditional in attribute grammar, e.g. Example 4. The second one is unique in ARSG. It helps guide the parsing process (shift or reduce).

\begin{myDef}
An attributed rhetorical structure grammar (ARSG) is a seven tuple:
$$(RS, DRE, DCP, RRE, PPR(AT), PF(AT), AT)$$
where $RS$ is the start symbol, $DRE$ the set of domain relations, $DCP$ the set of domain concepts, $RRE$ the set of rhetorical relations, $AT$ the synthesized attributes, where each attribute is an (arithmetical or logical) function with grammar symbols as arguments. $PPR(AT)$ is the set of probabilistic production rules, production rules for short, attached with attributes and reasons, where a reason is a propositional formula characterizing the state of parsing, e.g. the values of attributes and cues. A production rule of $PPR(AT)$ has the following form:
$$(ae): D(Lf) \gets A(X), B(Y); \ n$$
where $A$, $B$, $D$ are DRE, $X$, $Y$ are $role$ attributes (nucleus and/or satellite), $ae$ is set of attribute equations, $Lf$ is a reason, $n$ is a positive integer used for dynamically calculating the probability of taking this rule for reduction. The reasons $Lf$ are used for resolving reduce-reduce conflicts during parsing.

$PF(AT)$ is a set of precedence tuples where each tuple is in the form
\begin{equation}
(A, B, C, \prec, slf_{A, B, C}, p_s) \quad Or \quad (A, B, C, \succ, rlf_{A, B, C}, p_r)
\tag{101}
\end{equation}
where $ABC$ is a string of three neighboring grammar symbols during parsing. $slf_{A, B, C}$ (short for shift logic function) is a reason for shifting the parser over $C$, while $rlf_{A, B, C}$ (short for reduce logic function) is a reason for reducing $(A, B)$ to some DRE concept. The truth values of both $slf_{A, B, C}$ and $rlf_{A, B, C}$ depend on the attribute values of $A$, $B$ and $C$. $p_s$ and $p_r$ are probabilities with $p_s + p_r = 1$. Thus the quintuples (101) are used for resolving shift-reduce conflicts.
\hfill $\square$\par
\label{ARSGDef}
\end{myDef}

That $C$ is taken into consideration (see (101)) means ARSG is an `one symbol looking forward' grammar. So it is also called ARSG(1). In principle ARSG($n$) for $n$$>$$1$ can be defined. But due to its high complexity in grammar learning we give up its consideration.

\begin{myDef}
An  attributed rhetorical structure tree (ARTR) over a text D is an ARSG(1) parsing tree over D.
\hfill $\square$\par
\label{ARTRDef}
\end{myDef}

\begin{figure}
\includegraphics[width=0.7\textwidth]{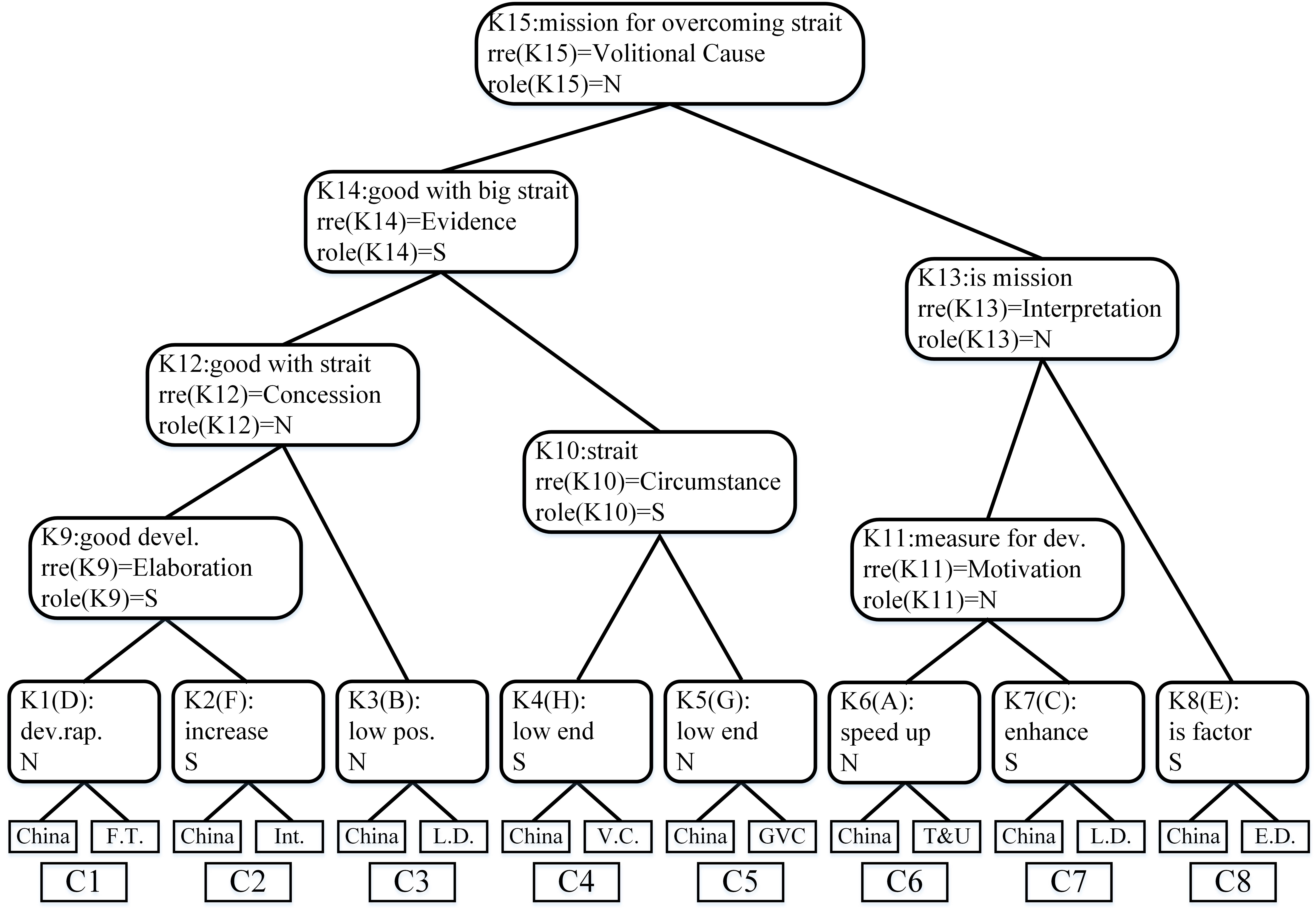}
\caption{An ARSG as parsing tree of Example~\ref{ftTextExamp}}
\label{fig5}
\end{figure}

\begin{myExamp}
Fig. \ref{fig5} shows an ARTR representing the parsing process of the text in Example~\ref{ftTextExamp}. On the bottom line, C1-C8 represent the eight clauses of the text.  K1-K8 are the roots of the eight basic trees which are the leaves of ARTR, where $i$-th basic tree is the tree representation of the LC of $i$-th clause C$i$. Above them, K9-K15 correspond to the non-leaf nodes of the parsing tree where each one is a domain relation (DRE). Among the attributes of DRE we only display the $rre$ attribute and $role$ attribute. For example, for the node K9, the domain relation is `good devel.', attribute $rre(K9)$ has value `Elaboration', while attribute $role(K9)$ has value `S' (i.e. satellite). We will see that attributes are very useful. They not only solve the representation problem, but also solve the parsing problem.

For the meaning of short notations in Fig. \ref{fig5} see the underlined parts of Example~\ref{GenExamp3GrammarExamp}.
\label{ARSTParsingProcessExamp}
\end{myExamp}

\section{Grammar Learning and Parser Implementation}
\label{GrammarLearning}
Now we are about to learn an ARSG from a set of mannually parsed domain specific NL texts. This is done by the following Procedure~\ref{learnFirstProcedure}-\ref{learnSecondProcedure} and Algorithm~\ref{grammarLearnAlgorithm}. Note that in our terminology, a procedure is a pseudo program performed by human programmer and computer cooperatively, while an algorithm is a program executed by  computer alone.

\begin{myProc}
Prepare the DKB, where H, M, A mean Human, Manually and Automatically resp.

1. (H) Determine a domain D of knowledge;

2. (M) Collect a number of e-dictionaries of domain D;

3. (A) Scan the e-dictionaries to extract all concepts from them;

4. (A) Use thesaurus~\cite{dongz2006hownetandthecomputationof,CheLL10} to enrich the concept set with their hypernyms, hyponyms and nicknames;

5. (A) Use these relations to construct three partial orderings of three types of concepts which form the DKB.
\hfill $\square$\par
\label{learnFirstProcedure}
\end{myProc}

Next we will describe how to learn an ARSG. Procedure~\ref{learnSecondProcedure} constructs a parsing tree from each sample text. Algorithm~\ref{grammarLearnAlgorithm} synthesizes all these parsing trees with statistic methods to generate the wanted ARSG. 

To learn an ARSG, the programmer should start from a training set of NL texts in the domain under consideration. He/She should simulate the parsing process of the future grammar manually to generate a parsing tree in form of ARTR from each text. The future ARSG is then constructed by statistically evaluating this set of ARTRs.

In this sense the next procedure is critical, where the computer should record each dynamical context which leads to the decision of the next parsing step, such as how the shift-reduce conflict was resolved, which logical condition (called reason) was used in each case, how the reduce-reduce conflict was resolved, which attribute equations were calculated and which attribute values were passed over in a bottom up way, etc. All these records will be used in the future parsing process. For convenience, we have implemented an annotation system based on Java Web techniques, which makes human programmer's simulation of the parsing process in an interactive way. All operations and context changes are recorded by the computer automatically. The programmer's work can therefore be reduced with the help of this system. 

Note that although ARSG is probabilistic, a single sample ARTR is not probabilistic. The probability is calculated by synthesizing the whole set of ARTRs. The following Procedure~\ref{learnSecondProcedure} constructs an ARTR in two stages. The first stage is preprocessing. It constructs all basic trees, including the first three steps of Procedure~\ref{learnSecondProcedure}. The second stage is (manual) parsing. It constructs the remaining part (main part) of ARTR, which is completed in step 4. This is why no production is produced in the first stage. All production rules $PPR(AT)$ and shift-reduce precedence rules $PF(AT)$ are generated in the second stage.

\begin{myProc}
Construct ARTRs (following Procedure~\ref{learnFirstProcedure}) and transform them to grammar component instances.

For each marked text T (i.e. text whose all LCs have been extracted) do the following:

1. (A) Scan the sentences and match their words against the DKB to find out all LCs (called $T'$).

2. (A) Turn each LC of $T'$ in a basic tree, with its blue concept $G_i$ as root and green resp. red concepts $A_i$ and $B_i$ as two leaves.

3. (M+A) Determine the attributes of the roots of all basic trees. Assign values to these attributes if possible. In particular, syntactic attributes like $cue$ and $punctuation$, and semantic attributes like $rre$, $role$, as well as positive and negative evaluations of concepts, should be assigned to $G_i$. (Note that syntactical attributes can be determined automatically, while some semantic ones should be decided manually. Therefore the mode is `M+A'.)

4. (M+A) If there is more than one tree then scan the roots of trees from left to right. Push the first two roots in a stack.

\textbf{Loop}: While there are at least two elements in the stack do (assume the top two are A, B):

\textbf{Begin} (Produce grammar component instances)
\begin{enumerate}[itemindent=2em, itemsep=0.1ex, label=(\alph*)]
\item If the programmer decides a shift action then the following is done automatically: push the next root (say $C$) into stack and produce the precedence tuple ($A, B, C, \prec, slf_{A, B, C}, 1$) where $slf_{A, B, C}$ is the reason of shift. Return.
\item If the programmer decides a reduce action with a new domain relation (say $D$) as target, then the following is done automatically: replace $A, B$ with $D$ as their parent node (a new root). Produce ($A, B, C, \succ, rlf_{A, B, C}, 1$) where $rlf_{A, B, C}$ is the reason for reduction, as well as a production instance
$$(ae): D (Lf) \gets A(X), B(Y))  \eqno{(102)}$$ 
where $Lf$ is the reason for taking the above rule for reduction, $ae$ is the set of attribute equations to be calculated during reduction, $X$ and $Y$ denote nucleus and/or satellite. In particular, all $cue$ attributes of $A$ and $B$ are uploaded to $D$ and become a part of $D$'s $cue$ attributes. Also the $role$ attributes $X$ and $Y$ are uploaded to D. Return.
\end{enumerate}

\textbf{End}

5. All ARTR are produced. Stop Procedure~\ref{learnSecondProcedure}.
\hfill $\square$\par 
\label{learnSecondProcedure}
\end{myProc}

\begin{myExamp}
Apply Procedure \ref{learnSecondProcedure} to Example~\ref{ftTextExamp} for generating grammar component instances.

1. After performing the first two steps, we get the sequence of LCs and corresponding basic trees:

\noindent
(LC1) \{\underline{China}, \underline{f}oreign \underline{t}rade, \underline{dev}elops \underline{rap}idly\}, (LC2) \{\underline{China}, \underline{int}egration into the global value chain, \underline{increas}ing\}, (LC3) \{\underline{China}, international \underline{d}ivision of \underline{l}abor, \underline{low}er \underline{pos}ition\}, (LC4) \{\underline{China}, \underline{g}lobal \underline{v}alue \underline{c}hain, \underline{low end}\}, (LC5) \{\underline{China}, \underline{g}lobal \underline{v}alue \underline{c}hain, \underline{low end}\}, (LC6) \{\underline{China}, \underline{t}ransformation and \underline{u}pgrading, \underline{speed up}\}, (LC7) \{\underline{China}, international \underline{d}ivision of \underline{l}abor, \underline{enhance}\}, (LC8) \{\underline{China}, \underline{e}conomic \underline{d}evelopment in the future, \underline{is} important \underline{factor}\}\footnote{The underlined words are labeled as the abbreviation of each LC in Fig. \ref{fig5}.}

\noindent
where most triples contain a green, red and blue concept. LC4 has borrowed a green and a blue concept from LC5, LC6 borrows a green concept from LC7. This is a technique used in ARSG to complement the imperfect clauses. The underlined parts correspond to short notations contained in Fig. \ref{fig5} above.

2. After performing step 3 we have the attributes with their values. Here we only list those of the first three basic trees:
\begin{eqnarray}
\nonumber
happy (K1)&=&1, cue (K1) = \{ `although' \};\\
\nonumber
happy (K2) &=&1, cue (K2) = empty;\\
\nonumber
happy (K3)&=&-1, cue (K3) = \{ `still' \}.
\end{eqnarray}

3. Step 4 of Procedure~\ref{learnSecondProcedure} is done level wise in a loop. We show the first few steps:
 
We use notation of Fig. \ref{fig5} to simplify the representation, where all nodes are numbered with letter $K$ followed by an integer. Assume the programmer decides to reduce nodes $K1$ and $K2$ to node $K9$, then node $K9$ and node $K3$ are reduced to node $K12$. These decisions lead to the building of following precedence tuples and production instances:
\begin{eqnarray}
\nonumber
(K1, K2, K3,&\succ&, rlf_{1, 2, 3}, 1);\\
\nonumber
(K9, K3, K4,&\succ&, rlf_{9, 3, 4}, 1);\\
\nonumber
 (K12, K4, K5,&\prec&, slf_{12, 4, 5}, 1);\\
\nonumber
(ae_{1, 2, 9}): K9 (Lf_{1, 2, 9}) &\gets& K1(N), K2(S);\\
\nonumber
(ae_{9, 3, 12}): K12 (Lf_{9, 3, 12}) &\gets& K9(S), K3(N).
\end{eqnarray}
where

\begin{flushleft}
$rlf_{1, 2, 3} = (happy(K1) > 0) \wedge (happy(K2) > 0) \wedge (happy(K3) < 0) \wedge (cue(K1) = \{`although' \}) \wedge (cue(K3) = \{`still' \})$;

$rlf_{9, 3, 4} = (happy(K9) > 0) \wedge (happy(K3) < 0) \wedge (cue(K9) = \{`although'\}) \wedge (cue(K3) = \{`still'\}) \wedge (punctuation(K3) = point)$;

$slf_{12, 4, 5} = (cue(K12) = \{`although', `still'\}) \wedge (punctuation(K12) = point) \wedge (cue(K4) = \{`especially'\}) \wedge (happy(K4) = 0)$;

$Lf_{1, 2, 9}=True; Lf_{9, 3, 12}=True$;

$ae_{1, 2, 9} = \{(cue(K9) = cue(K1) \cup cue(K2), happy(K9)=1, rre(K9)=Conjunction\}$;

$ae_{9, 3, 12} = \{(cue(K12) = cue(K9) \cup cue(K3), happy(K12)=-1, rre(K9)=Concession\}$.
\end{flushleft}

In the above representation, each reason is the conjunction of all attribute propositions  of that part of text under parsing.
\label{GenExamp3GrammarExamp}
\end{myExamp}

\begin{algorithm}[t]
\caption{(Following Procedure~\ref{learnSecondProcedure}) Assume the construction of ARTRs for all training texts is finished and grammar component instances generated. Learn an ARSG ($RS, DRE, DCP, RRE, PPR(AT), PF(AT), AT)$}
\label{grammarLearnAlgorithm}
\begin{algorithmic}[1] 
\Require 
The start symbol $RS$ and four sets $DRE$, $DCP$, $RRE$, $AT$ defined beforehand.
\Ensure 
$PPR(AT)$, $PF(AT)$.
\State Consider all the sets:
\begin{equation}
\begin{aligned}
\{ (A, B, C, \prec, slf_i, 1), 1 \le i \le n_{ABC} \} \quad \mbox{and} \quad \{ (A, B, C, \succ, rlf_j, 1), 1 \le j \le m_{ABC} \}
\end{aligned}
\tag{103}
\end{equation}
constructed in Procedure~\ref{learnSecondProcedure}, where $A$, $B$, $C$ are $DRE$.
For each above multi-set construct precedence tuples:
\begin{center}
$(A, B, C, \prec, slf, p_s)$ and/or $(A, B, C, \succ, rlf, p_r)$
\end{center}
where 
$$slf =  \displaystyle \mathop \vee \limits_i (slf_i), \quad p_s = n_{ABC} / (n_{ABC}+m_{ABC})$$
$$rlf =  \displaystyle \mathop \vee \limits_j (rlf_j), \quad p_r = m_{ABC} / (n_{ABC}+m_{ABC})$$

Let 
\begin{equation}
PF(AT) = \{ (A, B, C, \prec, slf, p_s) \mid if \ it \ exists\} \cup \{(A, B, C, \succ, rlf, p_r) \mid if \ it \ exists\}
\tag{104}
\end{equation}

where all precedence tuples with $p_s = 0$ or $p_r = 0$ are removed.

\State For each pair of $DRE$ concepts ($A$, $B$), investigate the multi-set of all rule instances generated in Procedure \ref{learnSecondProcedure} in the following form:
$$(ae_i): D_i(Lf_i)  \gets A(X_i), B(Y_i), \ 1 \le i \le G$$
which means if $Lf_i = true$ then the concept string $AB$ can be reduced to $D_i$, where $(X_i,Y_i)$ equals (N, S) or (S, N) or (N, N) with N as nucleus and S as satellite, $G$ is the total number of rule instances whose right sides are $(A, B)$. During reduction of the $i$-th rule the set of attribute equations $ae_i$ will be calculated and performed.

Classify the rules in $t$ sets of essentially identical copies, where rules in the same set are identical except the reasons $Lf_{ij}$ which are allowed to be different:
$$(ae_i): D_i(Lf_{ij})  \gets A(X_i), B(Y_i), 1 \le i \le t, 1 \le j \le j_i, \sum \limits_i j_i=G$$
Cluster each $i$-th set to a single rule with a count of copy numbers, where the reason $Lf_i$ is the disjunction of all $Lf_{ij}$:
\begin{equation}
\begin{aligned}
(ae_i): D_i(Lf_i)  \gets A(X_i), B(Y_i), \ j_i, \ 1 \le i \le t
\end{aligned}
\tag{105}
\end{equation}
where $j_i$ is now the weight of the $i$-th rule. Note that all components in the rules are depending on particular $A$ and $B$. We omit their indices when no ambiguity exists. The sets (105) for all possible $A$, $B$ are the learned $PPR (AT)$ of the grammar. 
\end{algorithmic} 
\end{algorithm}

Note that the number attached to each rule of (105) in Algorithm~\ref{grammarLearnAlgorithm} is a positive integer, not a probability. The probability is calculated at parsing time as follows:

When the programmer decides to reduce the string $AB$, the computer finds all rules with $A$, $B$ as right side, whose reason $Lf$ is implied by the current parsing context (current attribute values of $A$ and $B$), and ignores all other rules.

Assume the rules whose $Lf$ are satisfied by the current parsing context have the indices \{$k$\}.
$$(ae_k): D_k(Lf_k)  \gets A(X_k), B(Y_k), \ j_k, \ \{ k \} \subseteq \{1 \le i \le g \}$$
Then the probability that the $h$-th rule, $h \in \{k\}$, is used for reduction is $j_h/\sum \limits_k j_k$.

\begin{myExamp}
Apply Algorithm \ref{grammarLearnAlgorithm} for generating grammar rules and applying them in parsing.

Assume Procedure \ref{learnSecondProcedure} generates following 23 rule instances:
\begin{eqnarray}
\nonumber
(q=r):&D(u<s) \gets A(N), B(S); \ 3\\
\nonumber
(q=r+1):&D(v<s) \gets A(S), B(N); \ 1\\
\nonumber
(m=q-1):&E(u<s) \gets A(S), B(N); \ 4 \\
\nonumber
 (m=q-1):&E(v<s) \gets A(S), B(N); \ 1\\
\nonumber
(q=r+1):&F(v<s) \gets A(N), B(S); \ 5 \\
\nonumber
 (q=r+2):&F(u<s) \gets A(N), B(S); \ 9
\end{eqnarray}
where $m, q, r, s, u, v$ are attributes, the digits after each rule denote the number of identical copies of that rule instance. Combine the third and fourth rule because they are equal except the $Lf$ values:
\begin{equation}
\nonumber
(m=q-1): E((u<s) \vee (v<s)) \gets A(S), B(N); \ 5
\end{equation}

As a result, the $PPR(AT)$ of the corresponding grammar has 5 rules in total.

Assume the parser decides to reduce $AB$. If the current parsing context satisfies $u < s$ but not $v < s$ then the following rules are applicable:
\begin{eqnarray}
\nonumber
(q=r)&:&D(u<s) \gets A(N), B(S); \ 3 \\
\nonumber
(m=q-1)&:&E((u<s) \vee (v<s)) \gets A(S), B(N); \ 5\\
\nonumber
(q=r+2)&:&F(u<s) \gets A(N), B(S); \ 9
\end{eqnarray}
The probability that ($A$, $B$) are reduced to $D(q = r)$ is $3/17$, $E(m = q-1)$ is $5/17$, $F(q = r+2)$ is $9/17$.

If $v < s$ but not $u < s$ is satisfied, then the following rules are applicable:
\begin{eqnarray}
\nonumber
(q=r+1)&:&D(v<s) \gets A(S), B(N); \ 1 \\
\nonumber
(m=q-1)&:&E((u<s) \vee (v<s)) \gets A(S), B(N); \ 5\\
\nonumber
(q=r+1)&:&F(v<s) \gets A(N), B(S); \ 5
\end{eqnarray}
The probability that ($A$, $B$) are reduced to $D(q = r+1)$ is $1/11$, $E(m = q-1)$ is $5/11$, $F(q = r+1)$ is $5/11$.

If both $u< s$ and $v< s$ are satisfied then all rules are applicable. 
The possibility that ($A$, $B$) are reduced to $D(q = r)$ is $3/23$, $D(q = r+1)$ is $1/23$, $E(m = q-1)$ is $5/23$, $F(q = r+1)$ is $5/23$, $F(q = r+2)$ is $9/23$.
\label{GenGrammarAndParsingExamp}
\end{myExamp}

Sometimes we need auxiliary information from other constituents of the sentence, such as numbers (they often reveal some statistical data), naming entities (they often denote some key persons or institutions) and cues (they often remind the role of the sentence unit). We take all these, in particular the cues, in consideration.

Cue phrases can be used as a sufficiently accurate indicator of rhetorical relations~\cite{Marcu2000The}. Unfortunately, not all the text units have cue phrases. We can't recognize rhetorical relations purely depending on cue phrases. But we can combine cue phrases and ARSG relation precedence parsing to improve parsing accuracy. The way to do this is to define cues and other syntactic marks as attributes, as we have done in the above algorithms.

\begin{myDef}
Domain independent cues \cite{Marcu2000The} are those parts of a statement, which notify the readers about the existence, confirmation, negation or transformation of text topics.
\hfill $\square$\par
\label{cuesDef}
\end{myDef}

\begin{myExamp}
\emph{if, if...else, otherwise, announce, say, although, still, nevertheless, especially, in particular, because, therefore, while} and \emph{whereas} are all domain independent cues.
\end{myExamp}
The parsing process of a text is done in two stages. The first stage transforms the whole text into a sequence of LCs in form of basic trees with their corresponding initial attributes. The second stage is the real parsing process. The whole learned attributed rhetorical structure grammar and cues are taken into consideration for text parsing. The goal is to find an ARTR of the highest plausibility. 

Algorithm \ref{grammarParsingAlgorithm} is a rough sketch of ARSG relation precedence parsing. Here a fail of backtracking means either no rule is applicable or all applicable rules have failed. 

\begin{algorithm}[t]
\caption{ARSG relation precedence parsing (A rough sketch).}
\label{grammarParsingAlgorithm}
\begin{algorithmic}[1] 
\State Preprocessing: Extract one LC from each clause if possible and build a basic tree from each LC. The initial string to be parsed is then the string of basic tree roots which are domain relations.
\State Use bottom up probabilistic attributed relation precedence parsing technique to parse the initial string until the final parsing tree ARTR is generated.
\State Use precedence tuples to resolve any shift-reduce conflict.
\State If reduce is to be done, use rule reason (parameter of rule head) to select a group of rules as candidates.
\State If the candidates are not unique, use rule numbers to calculate probability for selecting a single rule to resolve the reduce-reduce conflict finally.
\State Apply the rule. Calculate all attribute equations.
\State If Step 3 or 4 fails, then try backtracking.
\State If all backtracking fails then stop the program. Parsing failed.
\end{algorithmic} 
\end{algorithm}

\section{ARSG-based Text Summarization}
\label{ARSGBasedTS}
Before further developing the idea of ARSG and discussing how to use it to analyze and summarize a NL text, we first present a technique of generating text summarization based on ARSG parsing.

Different from the majority of literature~\cite{Marcu2000The,Barzilay1997,SilberM02,ErcanC08}, our summarization technique is purely based on the generated ARTR rather than selecting most important EDUs from the text directly. Generally, a text reporting/commenting the circumstance of a domain consists of two parts: a review of the current situation and a proposal about what to do next. This corresponds to the two subtrees of the ARTR generated (Note that an ARTR is always binary). Since we want the generated summaries to be always balanced between review and proposal (no matter how many EDUs are extracted), the selection of EDUs should be alternated between these two subtrees whenever both subtrees are not empty.

When going to produce a summary, the summarizer traverses the ARTR in a nucleus preference way:
\begin{enumerate}
\item It starts from ARTR's N-subtree (whose root is a nucleus) and traverses ARTR's N- and S-subtree (whose root is a satellite) alternatively;
\item For any subtree T the traverse order is T's root, T's N-subtree, T's S-subtree (or T's another N-subtree if T has two N-subtrees);
\item Whenever a leaf node (LC) is traversed, the sentence unit represented by it will be outputted;
\item The traversal will be stopped if all nodes have been traversed or enough number of EDUs have been outputted.
\end{enumerate}

For implementing the balanced output of EDUs declared above, we apply the coroutine technique. A coroutine consists of a finite set of subroutines. Although each subroutine runs independent, different subroutines may run in an interleaving way. The switch instruction interrupts the running of a subroutine and switches the control to another subroutine. Later when the control is switched back to the original subroutine, the latter continues running from the interrupted point.

\begin{algorithm}[t]
\caption{Nucleus centered ARTR summary generation}
\label{nwalkOfTAlgorithm}
Given the ARTR $T$;
The number of text EDUs $h(h>0)$; 
The number of desired summary EDUs $m(0<m<h)$; 
The reduction rate of summary $t(0<t \le 1)$; 
The number of already outputted EDUs $i:=0$;

If $R(T)=T$ then output $R(T)$ and stop, else call coroutine N;  

coroutine N:  $n-traverse(nuc(R(T)))$;

coroutine S:  $s-traverse(sat (R(T)))$.

\textbf{Sub-algorithm}:

** $n-traverse(X)$ =
\{ if $X$ is a leaf node then \{output $X$, if $i=m$ or $i/h \ge t$ then stop, else $i:=i+1$, switch\} else \{$n-traverse(nuc(X))$, switch, $n-traverse(sat(X))$, switch\}\};

** $s-traverse(X)$ =
\{ if $X$ is a leaf node then \{output $X$, if $i=m$ or $i/h \ge t$ then stop, else $i:=i+1$, switch\} else \{$s-traverse(nuc(X))$, switch, $s-traverse(sat(X))$, switch\}\};

where $R(T)$ is the root of tree $T$, $nuc(X)$ and $sat(X)$ mean the nucleus resp. satellite child nodes of $X$. 
\end{algorithm}

\begin{mytheorem}  
Algorithm \ref{nwalkOfTAlgorithm} has the following properties:

1). It terminates for any finite ARTR;

2). It is ergodic in the sense that all nodes will be scanned (Therefore all EDUs corresponding to the leaf nodes will be outputted) unless it is terminated because enough EDUs have been outputted;

3). Each node will be scanned at most once (Therefore no EDU will be outputted twice);

4). EDUs corresponding to the left and right subtrees will be outputted alternatively;

5). EDUs corresponding to any subtree $T$ will be outputted in the ``nucleus first'' way, i.e. EDUs corresponding to T's N-subtree will be outputted earlier than those corresponding to T's S-subtree;

6). Therefore, for each of ARTR's two subtrees, the order of outputting EDUs is the lexicographic order of the paths leading to the leaf nodes, where nucleus is preferred over satellite. For example, the path `N-N-S' is preferred over `N-S-N';

7). No matter how long the outputted summary is, it is always balanced in the same sense as the original text's rhetorical structure.

8). The complexity of the algorithm is linear.
\hfill $\square$\par
\label{nwalkOfTAlgorithmTheorem}
\end{mytheorem} 

\begin{myProof}
For 1). Yes, because each call of the traverse function executes only a finite number of instructions and makes only a finite number of further calls. Each call of traverse function goes a level of the tree deeper. The height of the parsing tree is finite. Therefore the number of instructions executed is finite. 

For 2). Yes, because of the recursive structure of the algorithm, the structure of a finite binary tree and the program halting condition $i=m$ or $i/h \ge t$.

For 3). Yes, because of the coroutine structure and the recursive structure of the traverse sub-algorithm.

For 4). Yes, because of the recursive structure of the traverse sub-algorithm. 

For 5). Yes, because each call of the traverse function goes a level of the tree deeper and no backtracking is needed.

For 6). Yes, because in the sub-algorithm $traverse(X)$, the recursive call of $traverse(nuc(X))$ is always before that of $traverse(sat(X))$.

For 7). Yes, the complexity is $O(n)$, where $n$ is the number of tree nodes.\hfill $\square$\par
\end{myProof}

\section{Experimental Results}
\label{ExperimentalResults}

We have run Procedure~\ref{learnFirstProcedure} and constructed the DKB of WET, where the concepts were collected from some e-dictionaries including an English-Chinese World economy and trade dictionary together with the `Thesaurus of Synonym Words', (TCE for short)~\cite{CheLL10}. The current scale of DKB is shown in Table~\ref{scaleOfDKBTable}.

\begin{table}
\caption{The Current Scale of DKB of WET Domain}
\label{scaleOfDKBTable}
\begin{tabular}{llll}
\hline\noalign{\smallskip}
  & Green Concepts  & Red Concepts  & Blue Concepts\\
\noalign{\smallskip}\hline\noalign{\smallskip}
Height of Hierarchy  & 3  & 3  & 4\\

Number of Concepts  & 334  & 347  & 339\\
\noalign{\smallskip}\hline\noalign{\smallskip}
\end{tabular}
\end{table}

To evaluate our parsing method, we prepared the data set according to Procedure~\ref{learnFirstProcedure}-\ref{learnSecondProcedure}. In the domain of WET, we collected 180 Chinese texts from the official Web pages of China's Ministry of Commerce\footnote{http://www.mofcom.gov.cn/} as our experimental corpus. It contains 940 paragraphs and 2774 sentences. Under support of a computer aided context analyzing system two programmers simulated the parsing process of all the 180 texts according to  Procedure \ref{learnFirstProcedure} and Procedure~\ref{learnSecondProcedure}. The basic data in  Procedure~\ref{learnFirstProcedure} and Procedure~\ref{learnSecondProcedure} are listed in Table~\ref{parasOfCorpusTable}, where each EDU corresponds to a domain oriented LC.

\begin{table}
\caption{Basic data in Procedure \ref{learnFirstProcedure} and Procedure \ref{learnSecondProcedure} of Our Experimental Corpus}
\label{parasOfCorpusTable}
\begin{tabular}{ll}
\hline\noalign{\smallskip}
Knowledge domain & WET \\
\noalign{\smallskip}\hline\noalign{\smallskip}
Number of texts  & 180 \\
Average length of per text  & 742 \\
Average number of sentences per text & 16 \\
Average number of EDUs per text  & 22 \\
Average length of per EDU  & 30 \\
\noalign{\smallskip}\hline
\end{tabular}
\end{table}

After performing Procedure~\ref{learnFirstProcedure} and Procedure~\ref{learnSecondProcedure}, we counted the manually parsed 180 ARTRs. 
We also analyzed the reasons of disagreement between two different programmers and made some regulations, following which two programmers parsed the same 50 texts independently to measure the consistency. Our experiments have shown that two programmers marking the same 1758 text segment pairs from 50 Chinese texts have produced 313 mismatches. The rate of mismatches is 17.80\%.

In the above parsing process, the RRE was grounded in the framework of RST~\cite{Mann1988Rhetorical}, whose latest relation definitions are available at the website\footnote{http://www.sfu.ca/rst/01intro/definitions.html}, DRE was grounded in the type system of blue concepts(Definition~\ref{threeDictDef}).

In the following, we will give a snapshot of ARTRs and the ARSGs which are learned from them.

Each ARTR has a similar structure as shown in Fig. \ref{fig5}, the leaves of which (i.e. the basic trees) have blue concepts as roots as defined in Definition \ref{threeDictDef}. Due to the different lengths of texts the widths of ARTRs are ranging from 12 to 148 with the average width 44. The depths of ARTRs are ranging from 5 to 17 with the average 9.2.

For generating the set $AT$ we defined 11 classes of 16 synthesized attributes which are required to cover all the aspects of ARSG parsing. Among these attributes, $rre$ denotes the rhetorical relation between two sibling nodes, $role$ means nucleus or satellite, $cue$ has the usual meaning, $happy$ means `positive' or `negative' regard to a state or state change. In total we collected 140 cues as values of $cue$ attribute. 

Then we learned $PPR(AT)$ and $PF(AT)$ from all the 180 ARTRs. For $PPR(AT)$, there are 3788 production rule instances in form of (102). They are classified them into 1015 sets of essentially identical copies as in Example~\ref{GenGrammarAndParsingExamp}. With the method explained in Procedure~\ref{learnSecondProcedure}, they are further synthesized in the same number of production rules in form of (105). For $PF(AT)$ we got 6883 preference rule instances in form of (103), which are then synthesized in step 2 of Algorithm~\ref{grammarLearnAlgorithm} to 1081 preference rules in form of (104).

\begin{figure}
\includegraphics[width=.95\textwidth]{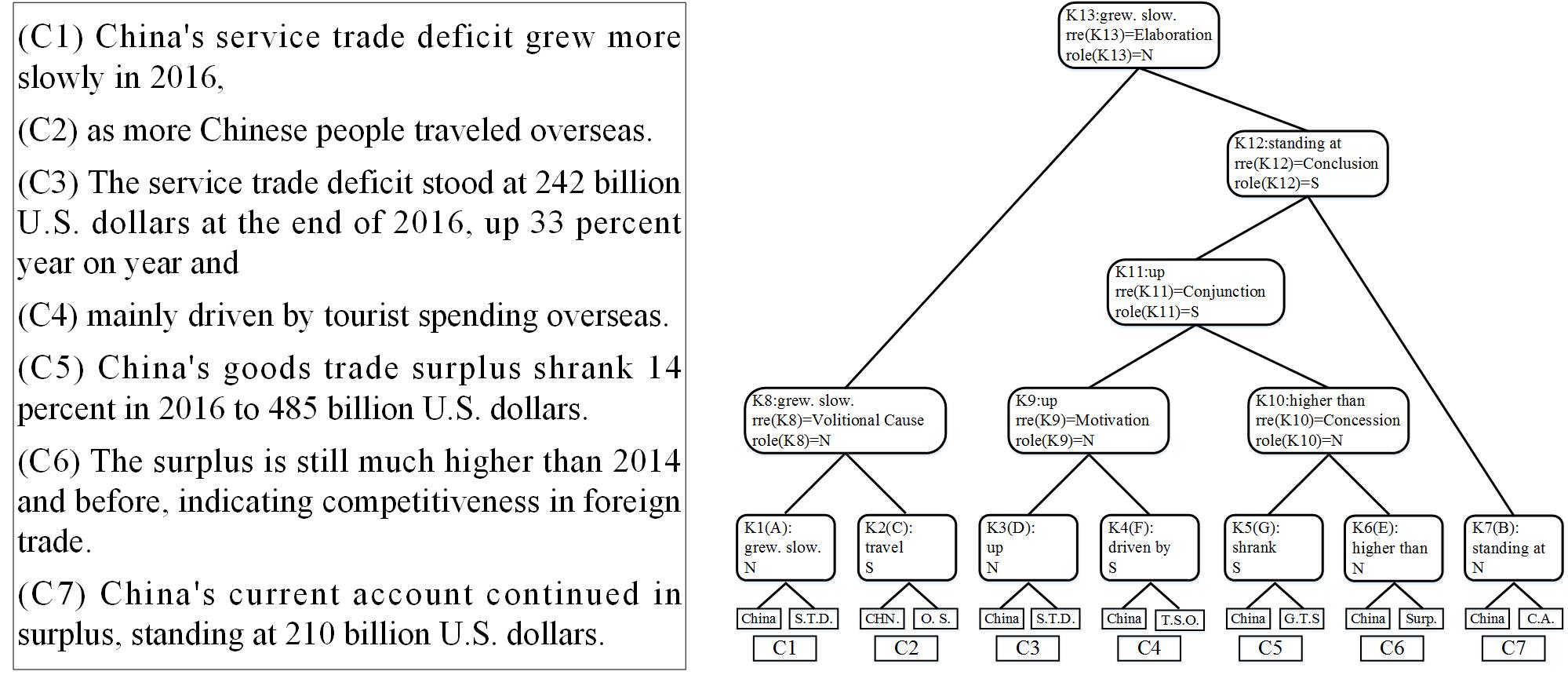}
\caption{A real case of our parsing method}
\label{Fig9_new_examp}
\end{figure}

\begin{myExprm}
Fig. \ref{Fig9_new_examp} is a real case of our parsing method. The input text on the left side is translated from Chinese. The ARTR on the right side is the output. Applying Algorithm~\ref{nwalkOfTAlgorithm} to the ARTR in Fig. \ref{Fig9_new_examp}, we get the sequence of outputs shown in Table~\ref{exampSumOutputSeqEaxmp}, where digital numbers denote the original order of the EDUs in the text, while capital letters denote the order of their generation by Algorithm~\ref{nwalkOfTAlgorithm}. In fact, this is also the order of their significance in the summary. In this way the users can request the summarizer to generate a summary of any length, which is always the most significant part of the original text. Each time the newly generated EDU will be inserted in the right place of the sequence of already outputted EDUs. For example, for $length$=1 one gets only one EDU ``A-1'', for $length$=2 one gets the EDU sequence (``A-1'', ``B-7''), etc. Note that in practical application, the redundant words will be deleted.
\label{ARTRInFig5SumExprm}
\end{myExprm}

\begin{table}
\caption{Layer-wise outputs of applying Alogrithm \ref{nwalkOfTAlgorithm} to ARTR in Fig. \ref{fig5}}
\label{sampleSummayOutTable}
\begin{tabular}{m{0.98\columnwidth}}
\hline\noalign{\smallskip}
A-1--China's service trade deficit grew more slowly in 2016, \\
C-2--as more Chinese people traveled overseas. \\
D-3--The service trade deficit stood at 242 billion U.S. dollars at the end of 2016, up 33 percent year on year and \\
F-4--mainly driven by tourist spending overseas. \\
G-5--China's goods trade surplus shrank 14 percent in 2016 to 485 billion U.S. dollars. \\
E-6--The surplus is still much higher than 2014 and before, indicating competitiveness in foreign trade. \\
B-7--China's current account continued in surplus, standing at 210 billion U.S. dollars.  \\
\noalign{\smallskip}\hline
\end{tabular}
\label{exampSumOutputSeqEaxmp}
\end{table}

\begin{figure}
\includegraphics[width=0.65\textwidth]{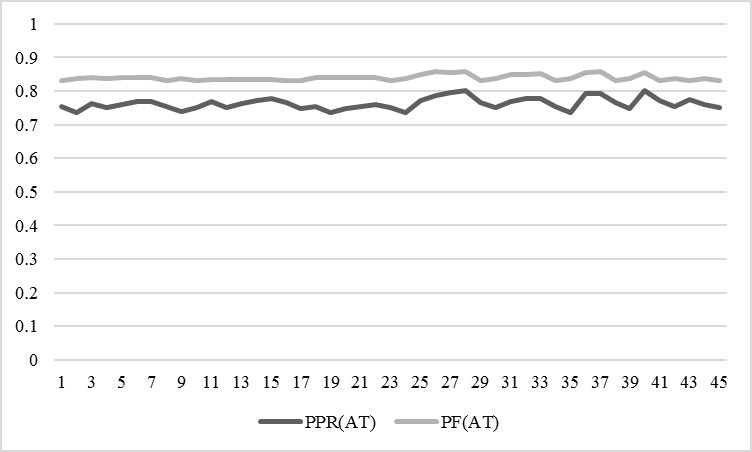}
\caption{Distribution of repetition rate of ten-fold cross-validation}
\label{fig6}
\end{figure}

To validate the effectiveness of grammar learning, we used ten-fold cross-validation to evaluate our parsing method, where each time 90\% of the data is used. Among the 10 learned ASRGs, the average size of $PPR (AT)$ and $PF(AT)$ are 948 and 1020 resp. Their average repetition rate is 76.28\% for $PPR(AT)$ and 83.97\% for $PF(AT)$, see Fig. \ref{fig6}.

We used precision, recall and F-score measurements to estimate the performance of our parsing algorithm (Algorithm~\ref{grammarParsingAlgorithm}). The estimation was based on comparing the performance of the machine learned ARSG parser with that of manually parsed ARTR. Precision reflects the rate of correctly parsed ARTRs among all machine parsed ones. Recall reflects the rate of correctly machine parsed ARTRs among all manually parsed ones. F-score is the harmonic mean of precision and recall. Note that the way of calculating precision, recall and F-score is the same for both manually constructed and machine parsed ARTR because they have the same number of leaf nodes~\cite{Marcu2000The,JotyCNM13}.

\begin{myExprm}[Text parsing using ARSG]

We used the 10 ARSGs for text parsing according to Algorithm~\ref{grammarParsingAlgorithm}. Note that when shift-reduce conflict or reduce-reduce conflict happens during parsing, the highest probability candidate production rule will be selected first while the second highest probability rule will be selected only if the first selection was not successful and backtracking is necessary.
\label{ARSGExprm}
\end{myExprm}

\begin{table}
\caption{Average correctness measured by ten-fold cross-validation}
\label{resultsOfARSGTable}
\begin{tabular}{lllll}
\hline\noalign{\smallskip}
Level & Structure & Nuclearity & RRE & DRE\\
\noalign{\smallskip}\hline\noalign{\smallskip}
Sentence-level & 98.76\% & 93.99\% & 87.80\% & 90.01\%\\
Paragraph-level & 93.55\% & 82.79\% & 71.28\% & 71.59\%\\
Discourse-level & 88.87\% & 75.01\% & 63.11\% & 61.36\%\\
\noalign{\smallskip}\hline
\end{tabular}
\end{table}

Table~\ref{resultsOfARSGTable} shows the mean correctness of 10-fold parsing using 10 ARSGs. As for the correctness of attribute use, we only evaluate those attributes on the correctly parsed nodes (e.g. correctly parsed DREs). Table~\ref{resultsOfAttrTable} shows the mean values, which illustrate that the mean precision is close to mean recall. The high F1-values indicate that almost all correctly parsed nodes have correct attributes.

\begin{table}
\caption{Correctness of attributes of correctly parsed nodes}
\label{resultsOfAttrTable}
\begin{tabular}{lll}
\hline\noalign{\smallskip}
Precision & Recall & F-score\\
\noalign{\smallskip}\hline\noalign{\smallskip}
92.23\% & 91.19\% & 91.71\%\\
\noalign{\smallskip}\hline
\end{tabular}
\end{table}

\begin{myExprm}[Evaluation of ARSG based summarization techniques used in parsing]

Our summarization technique, which is elaborated in Algorithm~\ref{nwalkOfTAlgorithm}, is purely based on the ARTR generated by ARSG parsing. As shown by Experiment 1, it can provide flexible and adjustable summaries either in form of an abstract in requested length or one according to requested reduction ratio. A synthetic experiment was made by learning an ARSG based on all 180 training texts. We applied it to parse text and produce the ARTR in Fig. \ref{Fig9_new_examp}.
\end{myExprm}

We also compared our method with other classical summarization methods, among which we used ROUGE~\cite{Flick2004ROUGE}, a recall-oriented summarization evaluation method, which measures the n-gram overlap between system generated and manually produced summaries, to evaluate these methods. The ROUGE scores are standard in automatic text summarization~\cite{ParveenM016}, among which we chose ROUGE-2 and ROUGE-S4 as our evaluation measures. Note that before the calculation of ROUGE scores, ICTCLAS Chinese word splitter~\cite{zhang2003ictclas} was used to split the summaries, where punctuations and stop words were excluded from matching.

We compared the performance of our ARSG method with those of ILP-based method~\cite{gillick2009scalable}, TextRank~\cite{Mihalcea2004TextRank}, Lead and Mead~\cite{RadevJST04} methods which are all classical summarization methods. ``ILP'' is a text summarization method which utilizes Integer Linear Programming (ILP) for inference under a maximum coverage model. ``TextRank'' is a graph-based summarization approach. It represents a text as a graph, whose nodes and edges correspond to sentences and their similarity. This model ranks sentences according to ``voting'' or ``recommendation'' from the adjacent sentences. ``Lead'' selects sentences from the beginning of an article. ``Mead'' computes summary sentences using cluster centroids produced by a topic detection and tracking(TDT) system.

\begin{table}
\caption{Rouge Scores(25\% Rate) of Comparative Evaluations}
\label{compareSumTable}
\begin{tabular}{lllllll}
\hline\noalign{\smallskip}
 & ILP & TextRank & Lead & Mead & ARSG \\
\noalign{\smallskip}\hline\noalign{\smallskip}
ROUGE-2 & 60.67\% & 59.44\% & 50.92\% & 56.93\% & 78.56\% \\
ROUGE-S4 & 54.82\% & 54.24\% & 46.38\% & 52.92\% & 77.63\% \\
\noalign{\smallskip}\hline
\end{tabular}
\end{table}

Table~\ref{compareSumTable} shows the evaluation results with reduce rate of 25\%. In the experiments on our collected Chinese corpus, the ROUGE scores of our method is better than the other methods. Our ``ARSG'' based summarization method enjoys the best performance with an improvement(ROUGE-2) of 17.89\% over ``ILP''.

\section{A Transductive Approach for Model Generalization}
\label{ModelTransduction}
The supervised learning of ARSG discussed above can be made more effective by introducing a transductive approach for model transfer. We assume that the different application domains (thus also their models) form a partial order according to the knowledge they contain. Among them, the domain of global situation analysis (GSA) is the root of a big tree whose nodes represent different child domains and thus also different child models. We do not need to construct a model for each node explicitly. By doing appropriate partial order operations it is possible to transfer models from one node to the others. By extending (reducing, transforming, crossing) knowledge bases and grammar rules we can upgrade (simplify, transfer, recombine) the models.  In this sense our models are transferable. The WET domain which has been taken as an example in this paper is just one of these nodes. It is not difficult to transfer WET oriented ARSG to other child domains of GSA.

\begin{myExprm}
 Experiment on other domain texts.
 
To validate the transferability of our method, we conduct an experiment on the domain: Industrial Dynamics (ID), which is another child domain of GSA and also a sister domain of WET.  To check the feasibility of transferring an ARSG model, we collected 100 Chinese texts from another open source: the official Web pages\footnote{http://www.miit.gov.cn/} of China's Ministry of Industry and Information Technology (MIIT). The average text length of this corpus is approximately equal to the corpus of WET.

\begin{table}
\caption{The Current Scale of DKB (After Extension) Used for ID Text Parsing}
\label{scaleOfIDDKBTable}
\begin{tabular}{llll}
\hline\noalign{\smallskip}
  & Green Concepts  & Red Concepts  & Blue Concepts\\
\noalign{\smallskip}\hline\noalign{\smallskip}
Number of Concepts  & 427  & 418  & 339\\
\noalign{\smallskip}\hline
\end{tabular}
\end{table}

\begin{table}
\caption{The Modification of ARSG from WET to ID}
\label{wetToIDModTable}
\begin{tabular}{llll}
\hline\noalign{\smallskip}
  & Production rules  & Attributes  & Precedence rules\\
\noalign{\smallskip}\hline\noalign{\smallskip}
Number of Changes  & 197  & 2  & 213\\
\noalign{\smallskip}\hline
\end{tabular}
\end{table}

 The experiment shows that with a slight extension (Table \ref{scaleOfIDDKBTable}) of the WET DKB and a simple transformation (Table \ref{wetToIDModTable}) of the ARSG rules it is possible to obtain a usable new ARSG for the ID domain. In fact, regarding the concept database, the blue ones are enough to be used for ID text parsing. Only a few new green concepts and red concepts should be added. Correspondingly, the amount of grammar rule modifications is shown in Table \ref{wetToIDModTable}. Comparing it with the ARSG learned above (1015 production rules, 1081 preference rules, see the statements before Experiment 1), the workload of generating an ARSG for the new domain ID is less than 1/5 of learning that ARSG for the original domain WET, not yet to mention the save of manual work on texts and manual construction of parsing trees for each text.
\label{IDExprm}
\end{myExprm}

We randomly selected 32 texts from this corpus, which are then parsed both mechanically by the transferred ARSG and manually by human. The results comparing between automatically generated ARTRs and manually parsed ARTRs are shown in Table \ref{resultsOfIDParsingTable}, which indicate the effectiveness of our method.

\begin{table}
\caption{Precision after Model Transfer}
\label{resultsOfIDParsingTable}
\begin{tabular}{m{0.2\columnwidth}<{\centering} m{0.1\columnwidth}<{\centering} m{0.12\columnwidth}<{\centering} m{0.1\columnwidth}<{\centering} m{0.1\columnwidth}<{\centering}}
\hline\noalign{\smallskip}
Level & Structure & Nuclearity & RRE & DRE\\
\noalign{\smallskip}\hline\noalign{\smallskip}
Sentence-level & 98.05\% & 91.27\% & 87.03\% & 89.26\%\\
Paragraph-level & 91.63\% & 74.22\% & 62.35\% & 63.96\%\\
Discourse-level & 83.13\% & 67.76\% & 51.26\% & 50.42\%\\
\noalign{\smallskip}\hline
\end{tabular}
\end{table}

\section{A Selective Comparative Study}
\label{ApproachComparation}

\begin{table}
\caption{Comparison of three Approaches}
\label{ComparisonOfTreeTable}
\begin{tabular}{m{0.17\columnwidth}<{\centering} m{0.2\columnwidth}<{\centering} m{0.21\columnwidth}<{\centering} m{0.22\columnwidth}<{\centering}}
\hline\noalign{\smallskip}
Approach  & Graph  & Neural Network  & ARSG\\
\noalign{\smallskip}\hline\noalign{\smallskip}
Representative work  & TextRank \cite{Mihalcea2004TextRank}  & NN-SE \cite{cheng2016neural}  & Our method\\
Paradigm & Unsupervised & Supervised & Supervised\\
Technique & Mainly Syntactical & Mainly Syntactical & Mainly Semantical\\
Information Source & Data Driven & Data Driven & Knowledge Driven\\
Data Request & Limited Data Set & Very Large Data Set & Limited Data Set\\
Knowledge Request & Few Linguistic Knowledge & Few Linguistic Knowledge & Domain Knowledge\\
Knowledge Benefiter & Human (Graph design) & Human (NN design) & Human(Training), Computer (Model construct \& parse)\\
Model Application & Domain Independent & Domain Independent & Domain Dependent\\
Design of Model Structure & Simple & Difficult, Need Try and Test & Fixed Once Designed\\
Training of Model Parameters & No Training & Algorithmic and Mechanically & Need Manual Training\\
Evolution of
Model Structure & Weighted Graph Revision & Heuristic Analysis and Revision of Model Behavior & Algorithmic--Incremental Evolution\\
Model Interpretability & Superficial & Poor & Professional\\
Generalization of Model & Simple reuse (result fair) & Simple reuse (result fair) & Simple reuse in the same domain (result good)\\
Applicability to Different Data Sources & Simple reuse (result fair), success rate with DUC2002 roughly equals to \cite{cheng2016neural} & Simple reuse (result fair), depending on the quality of training data (double success rate with DUC2002 than with DailyMail) & Simple reuse for new domain (result inadequate); Transfer to new domain need DKB extension and model mapping\\
Preciseness of Summary (ROUGE-2) & Generally low & Generally low & Generally high \\
\noalign{\smallskip}\hline
\end{tabular}
\end{table}

Table \ref{ComparisonOfTreeTable} compares our ARSG approach with the two most followed approaches: the graph based \cite{Mihalcea2004TextRank} resp. NN-based \cite{cheng2016neural} approaches. 

In addition, we also compare ARSG with ILP methods. Since the later need calculate relevance and redundancy weights of sentence resp. sentence pairs, we consider it as based on graph methods. Roughly speaking, ILP method can provide high quality summaries but has difficulty in processing large-sized documents since its complexity is NP-hard \cite{gillick2009scalable}.

\section{Concluding Remarks}
\label{ConcludingRemarks}
The contributions of this paper include:
1. Introduced an attribute grammar based approach to study automatic text summarization;
2. Introduced rhetorical structure theory in this approach to help NL analysis;
3. Combine the above two to form a framework of ARSG;
4. Proposed and implemented effective algorithms for machine learning an ARSG;
5. Proposed and implemented effective algorithm for parsing NL texts in ARSG framework;
6. Proposed and implemented RST guided algorithm for generating adjustable summaries;
7. Proposed a transductive approach for model transfer to avoid data retraining when given ARSG applied to new domains;
8. Performed a series of experiments to validate above results.

In order to save the work of building a new DKB for each new domain, we introduced a transduction approach for model transfer. Two different domains often have some similarities regarding the abstract representation of their inherit knowledge. Take the two example domains WET and ID considered in this paper. The concepts `output' of WET and `production' of ID are different. But the rule `if output increases then situation is improved' can be mapped to `if production increases then situation is improved' given `output' is mapped to `production'. This reminds that (afresh) retraining is not always necessary when transferring to a new domain. This is shown by Experiment \ref{IDExprm} and Tables \ref{scaleOfIDDKBTable}, \ref{wetToIDModTable} and \ref{resultsOfIDParsingTable}. In particular, Table \ref{wetToIDModTable} shows that the workload of obtaining an ID model by transferring is roughly only $1/5$ of that for building the WET model. 

It is also a new idea in this paper that we propose a method of combining the RST with the lexical chain technique, i.e. combining the coherent approach with the cohesive approach. These two approaches have been forming the main streams of text summarization research. Roughly speaking, the former is a top down technique and the latter is a bottom up one. If isolated from each other, both have their own advantages and disadvantages.  However, their combination will provide new vigor for both of them by augmenting their advantages and diminishing their disadvantages. This combination, of course, should not be a simple `put together'. Both sides should adapt themselves to meet the other side. We have changed lexical chains into lexical cores, on the one hand, and extended rhetorical relations with domain relations, on the other hand, this strategy has made our approach operational. 

In spite of the success made by the RST technique in summarization technique research, the RST based summarization technique is still in the phase of laboratory experiments. We have appreciated a large set of RST research papers with outstanding experimental results. But we didn't see its engineering, even less commercial, application yet.

Our future work will be concentrated on deepening and broadening the application range of ARSG approach, including: 
1. Applying the ARSG approach to a hierarchy of domains systematically and constructing a hierarchy of ARSG on it to further test its model transfer capability;
2. Introducing ARSG based summarization of multiple texts. Discussing the specific difficulties and techniques encountered in this context;
3. Making the above research results an engineering discipline and a public service for practical use;
4. Comparing ARSG based summarization techniques with other approaches such as deep learning techniques to form more efficient new approaches.

\begin{acknowledgements}
This work was supported by National Key Research and Development Program of China under grant 2016YFB1000902, National Natural Science Foundation of China (No. 61232015, 61472412, and 61621003), Beijing Science and Technology Project: Machine Learning based Stomatology and Tsinghua-Tencent-AMSS Joint Project: WWW Knowledge Structure and its Application.
\end{acknowledgements}

%
%

\bibliographystyle{spmpsci}      
\bibliography{ref}   

\end{document}